\documentclass{article}

\usepackage{microtype}
\usepackage{graphicx}
\usepackage{subfigure}
\usepackage{booktabs} % for professional tables

\usepackage[normalem]{ulem}
\usepackage{amsmath}
\usepackage{amssymb}
\usepackage{mathtools}

\DeclarePairedDelimiterX{\infdivx}[2]{(}{)}{%
  #1\;\delimsize\|\;#2%
}
\newcommand{\KL}{\text{KL}\infdivx}

\usepackage{bm}

\usepackage{multirow}

\usepackage{pgf}
\usepackage{tikz}
\usetikzlibrary{shapes,arrows,automata,backgrounds,calc}

\usepackage{array}
\newcolumntype{x}[1]{>{\centering\arraybackslash\hspace{0pt}}p{#1}}

\usepackage{colortbl}

\usepackage{hyperref}

\usepackage[accepted]{icml2018}

\icmltitlerunning{Probabilistic Recurrent State-Space Models}

\begin{document}

\twocolumn[

\icmltitle{Probabilistic Recurrent State-Space Models}

\begin{icmlauthorlist}
\icmlauthor{Andreas Doerr}{bosch,amd}
\icmlauthor{Christian Daniel}{bosch}
\icmlauthor{Martin Schiegg}{bosch}
\icmlauthor{Duy Nguyen-Tuong}{bosch}
\icmlauthor{Stefan Schaal}{amd,usc}
\icmlauthor{Marc Toussaint}{mlr}
\icmlauthor{Sebastian Trimpe}{amd}
\end{icmlauthorlist}

\icmlaffiliation{bosch}{Bosch Center for Artificial Intelligence, Renningen, Germany}
\icmlaffiliation{amd}{Max Planck Institute for Intelligent Systems, Stuttgart/T{\"u}bingen, Germany}
\icmlaffiliation{mlr}{Machine Learning and Robotics Lab, University of Stuttgart, Germany}
\icmlaffiliation{usc}{Computational Learning and Motor Control Lab, University of Southern California, USA}

\icmlcorrespondingauthor{Andreas Doerr}{andreasdoerr@gmx.net}

\icmlkeywords{time series, dynamics model, state-space model, Gaussian process, long-term prediction, variational inference}

\vskip 0.3in
]

\printAffiliationsNoNotice{}

\begin{abstract}
	State-space models (SSMs) are a highly expressive model class for learning patterns in time series data and for system identification.
Deterministic versions of SSMs (e.g. LSTMs) proved extremely successful in modeling complex time series data.
Fully probabilistic SSMs, however, are often found hard to train, even for smaller problems.
To overcome this limitation, we propose a novel model formulation and a scalable training algorithm based on doubly stochastic variational inference and Gaussian processes.
In contrast to existing work, the proposed variational approximation allows one to fully capture the latent state temporal correlations.
These correlations are the key to robust training.
The effectiveness of the proposed PR-SSM is evaluated on a set of real-world benchmark datasets in comparison to state-of-the-art probabilistic model learning methods.
Scalability and robustness are demonstrated on a high dimensional problem.
	\vspace{-5mm}
\end{abstract}

\section{Introduction}
\label{sec:Introduction}

System identification, i.e. learning dynamics models from data \cite{ljung1998system, ljung2010perspectives}, is a key ingredient of model-predictive control \cite{camacho2013model} and model-based reinforcement learning (RL) \cite{deisenroth2011pilco, doerr2017model}.
State-Space Models (SSMs) are one popular class of representations \cite{billings2013nonlinear}, which describe a system with input $\bm{u}_t$ and output $\bm{y}_t$ in terms of a latent Markovian state $\bm{x}_t$.
Based on a transition model $f$ and an observation model $g$, as well as process and measurement noise $\bm{\epsilon}_t$ and $\bm{\gamma}_t$, a time-discrete SSM is given by
\begin{align}
\bm{x}_{t+1} &= f(\bm{x}_t, \bm{u}_t) + \bm{\epsilon}_t\,, \nonumber \\
\bm{y}_{t} &= g(\bm{x}_t, \bm{u}_t) + \bm{\gamma}_t\,.
\label{eq:SSM}
\end{align}
Typically, in real systems, the latent state $\bm{x}_t$ cannot be measured directly but has to be inferred from a series of noisy output observations.
For \textit{linear} SSMs, this inference problem and the model learning can be solved simultaneously by subspace identification \cite{van2012subspace}.
Efficient methods also exist for the \textit{deterministic}, but \textit{non-linear} counterpart, e.g. recurrent neural networks (RNNs) \cite{hochreiter1997lstm}.

However, for tasks such as learning control policies, probabilistic models enable safe learning and alleviate model bias \cite{deisenroth2011pilco}.
Robust training of \textit{probabilistic, non-linear} SSMs is a challenging and only partially solved problem, especially for higher dimensional systems \cite{frigola2013bayesian, frigola2014variational, eleftheriadis2017identification, svensson2017flexible}.
This paper proposes the Probabilistic Recurrent State-Space Model \footnote{Code available after publication at: \url{https://github.com/andreasdoerr/PR-SSM}\,.} (PR-SSM\footnote{Pronounced \textit{prism}.}), a framework which tackles the key challenges preventing robust training of probabilistic, non-linear SSMs.
PR-SSM takes inspiration from RNN model learning.
In particular, the latent state transition model is unrolled over time, therefore accounting for temporal correlations whilst simultaneously allowing learning by backpropagation through time and mitigating the problem of latent state initialization.
Grounded in the theory of Gaussian Processes (GPs), the proposed method enables probabilistic model predictions, inferring complex latent state distributions, and principled model complexity regularization.
Furthermore, we propose an adapted form of a recognition model for the initial state distribution.
This facilitates scalability through batch learning and learning of slow or unstable system dynamics.

In summary, the key contributions of this paper are:
\vspace{-4mm}
\begin{itemize}
	\setlength\itemsep{-1mm}
	\item Combining gradient-based and sample-based inference for efficient learning of nonlinear Gaussian process state-space models.
	\item Tractable variational approximation, maintaining the true latent state posterior and temporal correlations.
	\item Doubly stochastic inference scheme for scalability.
	\item Recognition model, which allows robust training and prediction by initializing the latent state distribution.
\end{itemize}
\vspace{-4mm}
Together, these contributions allow for robust training of the PR-SSM.
The proposed framework is evaluated on a set of real-world system identification datasets and benchmarked against a range of state-of-the art methods.

\section{Related Work}
\label{sec:RelatedWork}

Modeling the behavior of systems with only partially observable states has been an active field of research for many years and several schools of thought have emerged.
Representations range from SSMs \cite{van2012subspace} over Predictive State Representations (PSRs) \cite{littman2002predictive, singh2003learning, rudary2004nonlinear} to autoregressive models \cite{murray2001gaussian, girard2003multiple, likar2007predictive, billings2013nonlinear}, as well as hybrid versions combining these approaches \cite{mattos2015recurrent, mattos2016latent, doerr2017optimizing}.

Autoregressive (history-based) methods avoid the complex inference of a latent state and instead directly learn a mapping from a history of $h$ past inputs and observations to the next observation, i.e. $\bm{y}_{t+1} = f(\bm{y}_{t:t-h}, \bm{u}_{t:t-h})$.
These models face the issue of feeding back observation noise into the dynamics model.
Recent work addresses this problem by either actively accounting for input noise \cite{mchutchon2011gaussian} or reverting to a hybrid, autoregressive formulation in a latent, but noise free state \cite{mattos2016latent, doerr2017optimizing}.
Such models can be made deep and trained in a recurrent manner as presented in \cite{mattos2015recurrent}.
In theory, a horizon $h$ identical to the true latent state dimensionality (dimension of $\bm{x}_t$) is sufficient to model all relevant dependencies of the system under consideration \cite{ljung1998system}.
In practice, however, autoregressive models typically need a much larger history horizon to cope with noisy observations and arbitrary sampling frequencies.

Thus, in this paper, we focus on SSMs based on a compact, Markovian state representations.
Furthermore, SSMs allow the direct application of many existing control algorithms, which rely on the explicit representation of the latent state.
Within the field of latent state models, exact solutions for state inference and model learning are known for linear SSMs and can be obtained by the well known Kalman filter/smoother \cite{kalman1960new} and subspace identification \cite{van2012subspace}.
In the case of non-linear latent state transition dynamics, both deterministic and probabilistic variants are active fields of research.

Deterministic variants such as Long Short-Term Memory (LSTM) models have been shown to be powerful representations for tasks such as natural language processing \cite{venugopalan2014translating} or text understanding \cite{sutskever2011generating}.
However, for the purpose of system identification and control, probabilistic predictions are often required to make model errors explicit \cite{deisenroth2011pilco}.
PR-SSM learning can be interpreted as a probabilistic version of the learning procedure in these deep recurrent models.
Both procedures share the explicit unrolling of transition and observation model.
Errors between the predicted and the observed system output are propagated back over time.
Therefore, the transition dynamics is to be learned but the latent state (distribution) is implicitly given.
This way, the challenging initialization and optimization of latent state variables is prevented.
In contrast to deep recurrent models, the PR-SSM loss and model regularization is automatically obtained from the model.
Furthermore, PR-SSMs obtain predictive distributions and the proposed initial state recognition model facilitates learning on shorter sub-trajectories and unstable systems, which is not possible in deep recurrent models.

Gaussian Process State-Space Models (GP-SSMs) are a popular class of probabilistic SSMs \cite{wang2008gaussian, ko2009gp, turner2010state, frigola2013bayesian, frigola2014variational, eleftheriadis2017identification}. 
The use of GPs allows for a fully Bayesian treatment of the modeling problem resulting in an automatic complexity trade-off, which regularizes the learning problem. 
Filtering and smoothing in GP-SSMs, has already been covered extensively: deterministic (e.g. linearization) as well as stochastic (e.g. particles) methods are presented in \cite{ko2009gp, deisenroth2012robust}.
These methods, however, assume an established system model which is generally not available without prior knowledge.
In this work, the latent state smoothing distribution is given implicitly and optimized jointly during model learning.

Approaches to probabilistic GP-SSMs mainly differ in their approximations to the model's joint distribution (e.g. when solving for the smoothing distribution or for the observation likelihood).
One class of approaches aims to solve for the true distribution which requires sample-based methods, e.g. Particle Markov Chain Monte Carlo (PMCMC), as in \cite{frigola2013bayesian, frigola2014variational}.
These methods are close to exact but computationally inefficient and intractable for higher latent state dimensions or larger datasets.
A second class of approaches is based on variational inference and mean field approximations \cite{mattos2015recurrent, foll2017deep}.
These methods, however, operate on latent autoregressive models which can be initialized by the observed output time series, such that the learned latent representation acts as a smoothed version of the observations.
In Markovian latent spaces, no such prior information is available and therefore initialization is non-trivial.
Model optimization based on mean field approximations empirically leads to highly suboptimal local solutions.
Bridging the gap between both classes, recent methods strive to recover (temporal) latent state structure.
In \cite{eleftheriadis2017identification}, a linear, time-varying latent state structure is enforced as a tractable compromise between the true non-linear dependencies and no dependencies as in mean field variational inference.
However, to facilitate learning, a complex recognition model over the linear time-varying dynamics is required.
In contrast, the proposed PR-SSM can efficiently incorporate the true dynamics by combining sampling- and gradient-based learning.
\section{Gaussian Process State-Space Model}
\label{sec:GaussianProcessStateSpaceModel}

This section presents the general model background for GP-SSMs.
Following a short recap of GPs in Sec.\,\ref{sec:GaussianProcess} and a specific sparse GP prior in Sec.\,\ref{sec:InducingGPTargets}, PR-SSM as one particular GP-SSM is introduced in Sec.\,\ref{sec:ModelDefinition}.

\subsection{Gaussian Process}
\label{sec:GaussianProcess}

A GP \cite{williams2006gaussian} is a distribution over functions $f:\mathbb{R}^D \rightarrow \mathbb{R}$ that is fully defined by a mean function $m(\cdot)$ and covariance function $k(\cdot, \cdot)$.
For each finite set of points $\bm{X} = [\bm{x}_1, \ldots, \bm{x}_N]$ from the function's domain, the corresponding function evaluations $\bm{f} = [f(\bm{x}_1), \ldots, f(\bm{x}_N)]$ are jointly Gaussian as given by
\begin{equation}
p(\bm{f} \mid \bm{X}) = \mathcal{N}(\bm{f} \mid \bm{m}_X, \bm{K}_{X,X})\,,
\label{eq:GPPrior}
\end{equation}
with mean vector $\bm{m}_X$ having elements $m_i = m(\bm{x}_i)$ and covariance matrix $\bm{K}_{X,X}$ with entries $K_{ij} = k(\bm{x}_i, \bm{x}_j)$.
Given observed function values $\bm{f}$ at input locations $\bm{X}$, the GP predictive distribution at a new input location $\bm{x}^*$ is obtained as the conditional distribution
\begin{equation}
p(f^* \mid \bm{x}^*, \bm{f}, \bm{X}) = \mathcal{N}(f^* \mid \mu, \sigma^2),
\label{eq:GPConditional}
\end{equation}
with posterior mean and variance
\begin{align}
\mu &= m_{x^*} + \bm{k}_{x^*, X} \bm{K}_{X,X}^{-1} (\bm{f} - \bm{m}_X)\,,
\label{eq:GPConditionalMean}\\
\sigma^2 &= k_{x^*, x^*} - \bm{k}_{x^*, X} \bm{K}_{X,X}^{-1} \bm{k}_{X, x^*}\,,
\label{eq:GPConditionalVariance}
\end{align}
where $k_{A, B}$ denotes the scalar or vector of covariances for each pair of elements in $\bm{A}$ and $\bm{B}$.
In this work, the squared exponential kernel with Automatic Relevance Determination (ARD) \cite{williams2006gaussian} with hyper-parameters $\theta_\text{GP}$ is employed.
Due to the proposed sampling-based inference scheme (cf.\,Sec.\,\ref{sec:DoublyStochasticVariationalInference}), any other differentiable kernel might be incorporated instead.

\subsection{GP Sparsification}
\label{sec:InducingGPTargets}

Commonly, the GP prediction in \eqref{eq:GPConditional} is obtained by conditioning on all training data $\bm{X}$, $\bm{y}$.
To alleviate the computational cost, several sparse approximations have been presented \cite{snelson2006sparse}.
By introducing $P$ inducing GP targets $\bm{z} = [z_1, \ldots, z_P]$ at pseudo input points $\bm{\zeta} = [\bm{\zeta}_1, \ldots, \bm{\zeta}_P]$, which are jointly Gaussian with the latent function $f$, the true GP predictive distribution is approximated by conditioning only on this set of inducing points,
\begin{align}
p(f^* \mid \bm{x}^*, \bm{f}, \bm{X}) &\approx p(f^* \mid \bm{x}^*, \bm{z}, \bm{\zeta})\,,
\label{eq:InducingGPPrediction} \\
p(\bm{z}) &= \mathcal{N}(\bm{z} \mid \bm{m}_\zeta, \bm{K}_{\zeta, \zeta})\,.
\label{eq:InducingGPPrior}
\end{align}
The predicted function values consequently become mutually independent given the inducing points.

\subsection{PR-SSM Model Definition}
\label{sec:ModelDefinition}

The PR-SSM is built upon a GP prior on the transition function $f(\cdot)$ and a parametric observation model $g(\cdot)$.
This is a common model structure, which can be assumed without loss of generality over \eqref{eq:SSM}, since any observation model can be absorbed into a sufficiently large latent state \cite{frigola2015bayesian}.
Eliminating the non-parametric observation model, however, mitigates the problem of \emph{`severe non-identifiability'} between transition model $f(\cdot)$ and observation model $g(\cdot)$ \cite{frigola2014variational}.
Independent GP priors are employed for each latent state dimension $d$ given individual inducing points $\bm{\zeta}_d$ and $\bm{z}_d$.

In the following derivations, the system's latent state, input and output at time $t$ are denoted by $\bm{x}_t \in \mathbb{R}^{D_x}$, $\bm{u}_t \in \mathbb{R}^{D_u}$, and $\bm{y}_t \in \mathbb{R}^{D_y}$, respectively.
The shorthand $\hat{\bm{x}}_t = (\bm{x}_t, \bm{u}_t)$ denotes the transition model's input at time $t$.
The output of the transition model is denoted by $\bm{f}_{t+1} = f(\hat{\bm{x}}_t)$.
A time series of observations from time $a$ to time $b$ (including) is abbreviated by $\bm{y}_{a:b}$ (analogously for the other model variables).

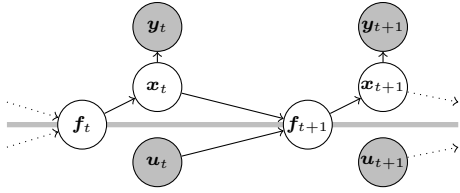
\begin{figure}
\vskip 0.2in
\begin{center}
\centerline{
	\begin{tikzpicture}[font=\fontsize{7}{7}\selectfont]
		\tikzstyle{every state}=[fill=white,circle,draw,minimum size=0.65cm,inner sep=0pt]
		\draw[line width=2, color=lightgray] (1,0.5) -- (7,0.5);
		\node[state] (Xt) at(3,1)        {$\bm{x}_t$};
		\node[state,fill=lightgray] (Yt) at(3,1.8)        {$\bm{y}_t$};
		\node[state,fill=lightgray] (Ut) at(3,0)        {$\bm{u}_t$};
		\node[state] (Xt1) at(6,1)        {$\bm{x}_{t+1}$};
		\node[state,fill=lightgray] (Yt1) at(6,1.8)        {$\bm{y}_{t+1}$};
		\node[state,fill=lightgray] (Ut1) at(6,0)        {$\bm{u}_{t+1}$};
		\node[state] (Ft) at(2,0.5)        {$\bm{f}_t$};
		\node[state] (Ft1) at(5,0.5)        {$\bm{f}_{t+1}$};
		\path[black,->] (Xt) edge (Yt);
		\path[black,->] (Ut) edge (Ft1);
		\path[black,->] (Xt) edge (Ft1);
		\path[black,->] (Ft1) edge (Xt1);
		\path[black,->] (Xt1) edge (Yt1);
		\path[black,->] (Ft) edge (Xt);
		\path[black,dotted,->] (1,0.2) edge (Ft);
		\path[black,dotted,->] (1,0.8) edge (Ft);
		\path[black,dotted,->] (Xt1) edge (7,0.8);
		\path[black,dotted,->] (Ut1) edge (7,0.2);
	\end{tikzpicture}
}
\caption{Graphical model of the PR-SSM. Gray nodes are observed variables in contrast to latent variables in white nodes. Thick lines indicate variables, which are jointly Gaussian under a GP prior.}
\label{fig:GraphicalModel}
\end{center}
\vskip -0.4in
\end{figure}

The joint distribution of all PR-SSM random variables is given by
\begin{align}
\label{eq:JointDistribution}
p(\bm{y}_{1:T}, \bm{x}_{1:T}, \bm{f}_{2:T}, \bm{z}) = 
& \left[ \prod_{t=1}^T p(\bm{y}_t \mid \bm{x}_t) \right] p(\bm{x}_1) p(\bm{z}) \\
& \left[ \prod_{t=2}^T p(\bm{x}_t \mid \bm{f}_t) p(\bm{f}_t \mid \hat{\bm{x}}_{t-1}, \bm{z})\right],    \nonumber
\end{align}
where $p(\bm{f}_t \mid \hat{\bm{x}}_{t-1}, \bm{z}) = \prod_{d=1}^{D_x} p(f_{t,d} \mid \hat{\bm{x}}_{t-1}, \bm{z}_d)$ and $\bm{z} \equiv [\bm{z}_1, \ldots \bm{z}_{D_x}]$.
A graphical model of the resulting PR-SSM is shown in Fig.\,\ref{fig:GraphicalModel}.

The individual contributions to \eqref{eq:JointDistribution} are given by the observation model and the transition model, which are now described in detail.
The observation model is governed by
\begin{equation}
p(\bm{y}_t \mid \bm{x}_t) = \mathcal{N}(\bm{y}_t \mid g(\bm{x}_t), \text{diag}(\sigma_{y,1}^2, \ldots, \sigma_{y, D_y}^2)),
\label{eq:ObservationModel}
\end{equation}
with observation function
\begin{equation}
g(\bm{x}_t) = \bm{C} \bm{x}_t\,.
\label{eq:ParametricObservationModel}
\end{equation}
In particular, the matrix $\bm{C}$ is chosen to select the $D_y$ first entries of $\bm{x}_t$ by defining $\bm{C} := \left[\bm{I}, \bm{0}\right] \in \mathbb{R}^{D_y \times D_x}$ with $\bm{I}$ being the identity matrix.
This model is suitable for observation spaces that are low-dimensional compared to the latent state dimensionality, i.e. $D_y < D_x$, which is often the case for physical systems with a restricted number of sensors.
The first $D_y$ latent state dimensions can therefore be interpreted as noise free sensor measurements.
For high-dimensional observation spaces (e.g.\,images), a more involved observation model (e.g.\,a neural network) may be seamlessly incorporated into the presented framework as long as $g(\cdot)$ is differentiable.
Process noise is modeled as
\begin{equation}
p(\bm{x}_t \mid \bm{f}_t) = \mathcal{N}(\bm{x}_t \mid \bm{f}_t, \text{diag}(\sigma_{x,1}^2, \ldots, \sigma_{x,D_x}^2))\,.
\label{eq:ProcessNoise}
\end{equation}
The transition dynamics is described independently for each latent state dimension $d$ by $p(f_{t, d} \mid \hat{\bm{x}}_{t-1}, \bm{z}_d) p(\bm{z}_d)$.
This probability is given by the sparse GP prior \eqref{eq:InducingGPPrior} and predictive distribution \eqref{eq:InducingGPPrediction}, where $x^* = \bm{\hat{x}}_t$ and $f^* = f_{t,d}$.
The initial system state distribution $p(\bm{x}_1)$ is unknown and has to be estimated. 

\section{PR-SSM Inference}
\label{sec:DoublyStochasticVariationalInference}

Computing the log likelihood or a posterior based on \eqref{eq:JointDistribution} is generally intractable due to the nonlinear GP dynamics model in the latent state.
However, the log marginal likelihood $\log p(\bm{y}_{1:T})$ (evidence) can be bounded from below by the Evidence Lower BOound (ELBO) \cite{blei2017variational}.
This ELBO is derived via Jensen's inequality by introducing a computationally simpler, variational distribution $q(\bm{x}_{1:T}, \bm{f}_{2:T}, \bm{z})$ to approximate the model's true posterior distribution $p(\bm{x}_{1:T}, \bm{f}_{2:T}, \bm{z} \mid \bm{y}_{1:T})$ (cf.\,eq.\,\eqref{eq:JointDistribution}).
In contrast to previous work \cite{frigola2014variational, mattos2015recurrent, eleftheriadis2017identification}, the proposed approximation explicitly incorporates the true temporal correlations in the latent state, whilst being scalable to large datasets.
The inference scheme is inspired by doubly stochastic variational inference for deep GPs as presented in \cite{salimbeni2017doubly}.

\subsection{Variational Sparse GP}
\label{sec:SparseGPApproximation}

PR-SSM employs a variational sparse GP \cite{titsias2009variational} based on a variational distribution $q(\bm{z})$ on the GP's inducing outputs as previously used in \cite{frigola2014variational, eleftheriadis2017identification}.
For the standard regression problem, the inducing output distribution can be optimally eliminated and turns out to be a Gaussian distribution.
Eliminating the inducing outputs, however, results in dependencies between inducing outputs and data which, in turn, leads to a complexity of $\mathcal{O}(NP^2)$, where $N$ is the number of data points and $P$ the number of inducing points \cite{titsias2009variational}.
Unfortunately, this complexity is still prohibitive for large datasets.
Therefore, we resort to an explicit representation of the variational distribution over inducing outputs as previously proposed in \cite{hensman2013gaussian}.
This explicit representation enables scalability by utilizing stochastic gradient-based optimization since individual GP predictions become independent given the explicit inducing points.
Following a mean-field variational approximation the inducing output distribution is given as $q(\bm{z}) = \prod_{d=1}^{D_x} \mathcal{N}(\bm{z}_d \mid \mu_d, \Sigma_d)$ for each latent state dimension $d$ with diagonal variance $\Sigma_d$.
Marginalizing out the inducing outputs, the GP predictive distribution is obtained as Gaussian with mean and variance given by
\begin{align}
\mu &= m_{\hat{\bm{x}}_t} + \bm{\alpha}(\hat{\bm{x}}_t) (\mu_d - m_{\bm{\zeta}_d})\,,\\
\sigma^2 &= k_{\hat{\bm{x}}_t, \hat{\bm{x}}_t} - \bm{\alpha}(\hat{\bm{x}}_t) (K_{\bm{\zeta}_d, \bm{\zeta}_d} - \Sigma_d) \bm{\alpha}(\hat{\bm{x}}_t)^T\,,\\
& \quad \bm{\alpha}(\hat{\bm{x}}_t) \coloneqq k_{\hat{\bm{x}}_t, \bm{\zeta}_d} K_{\bm{\zeta}_d, \bm{\zeta}_d}^{-1}\,.
\label{eq:GPPosteriorDetail}
\end{align}

\subsection{Variational Approximation}
\label{sec:VariationalApproximation}

In previous work \cite{mattos2015recurrent}, a factorized variational distribution is considered based on a mean-field approximation for the latent states $\bm{x}_{1:T}$.
Their variational distribution is given by
\begin{align}
q(\bm{x}_{1:T}, \bm{f}_{2:T}, \bm{z}) = &\left[ \prod_{d=1}^{D_x} q(\bm{z}_d) \left[\prod_{t=2}^T  p(f_{t,d} \mid \hat{\bm{x}}_{t-1}, \bm{z}_d) \right] \right] \nonumber \\
&\left[ \prod_{t=1}^T q(\bm{x}_t) \right] \nonumber \,.
\end{align}
This choice, however, leads to several caveats:
(i) The number of model parameters grows linearly with the length of the time series since each latent state is parametrized by its individual distribution $q(\bm{x}_t)$ for every time step.
(ii) Initializing the latent state is non-trivial since the observation mapping is unknown and generally not bijective.
(iii) The model design does not represent correlations between time steps.
Instead, these correlations are only introduced by enforcing pairwise couplings during the optimization process.
The first two problems have been addressed in \cite{mattos2015recurrent, eleftheriadis2017identification} by introducing a \textit{recognition} model, e.g. a Bi-RNN\footnote{A bi-directional RNN operates on a sequence from left to right and vice versa to obtain predictions based on past and future inputs.}, which acts as a smoother which can be learned through backpropagation and which allows to obtain the latent states given the input/output sequence.

The issue of representing correlations between time steps, however, is currently an open problem which we aim to address with our proposed model structure.
Properly representing these correlations is a crucial step in making the optimization problem tractable in order to learn GP-SSMs for complex systems.

For PR-SSM, the variational distribution is given by
\begin{align}
\label{eq:VariationalDistribution}
q(\bm{x}_{1:T}, \bm{f}_{2:T}, \bm{z}) = 
&\left[ \prod_{t=2}^T p(\bm{x}_t \mid \bm{f}_t) \right] q(\bm{x}_1) \cdot \\
&\left[ \prod_{t=2}^T \prod_{d=1}^{D_x} p(f_{t, d} \mid \hat{\bm{x}}_{t-1}, \bm{z}_d) q(\bm{z}_d) \right]\,, \nonumber
\end{align}
with 
\begin{equation}
q(\bm{x}_1)\!=\!\mathcal{N}(\bm{x}_1 \mid \bm{\mu}_{x_1}, \bm{\Sigma}_{x_1})\,, q(\bm{z}_d)\!=\!\mathcal{N}(\bm{z}_d \mid \bm{\mu}_d, \bm{\Sigma}_d)\,. \nonumber
\end{equation}

In contrast to previous work, the proposed variational distribution does not factorize over the latent state but takes into account the true transition model, based on the sparse GP approximation from \eqref{eq:JointDistribution}.
In previous work, stronger approximations have been required to achieve an analytically tractable ELBO.
This work, however, deals with the more complex distribution by combining sampling and gradient-based methods.

In \cite{frigola2014variational}, the variational distribution over inducing outputs has been optimally eliminated.
This leads to a smoothing problem in a second system requiring computationally expensive, e.g. sample-based, smoothing methods.
Instead, we approximate the distribution by a Gaussian, which is the optimal solution in case of sparse GP regression (cf.\,\cite{titsias2009variational}).

The PR-SSM model parameters include the variational parameters for the initial state and inducing points as well as deterministic parameters for noise models and GP kernel hyper-parameters:
$\theta_\text{PR-SSM} = (\bm{\mu}_{x_1}, \bm{\Sigma}_{x_1}, \bm{\mu}_{1:D_x}, \bm{\Sigma}_{1:D_x}, \bm{\zeta}_{1:D_x}, \sigma^2_{\text{x}, {1:D_x}}, \sigma^2_{\text{y}, {1:D_y}}, \theta_{\text{GP}, {1:D_x}})$.
Note that in the PR-SSM, the number of parameters grows only with the number of latent dimensions, but not with the length of the time series.

\subsection{Variational Evidence Lower Bound}
\label{sec:VariationalEvidenceLowerBound}

Following standard variational inference techniques \cite{blei2017variational}, the ELBO is given by
\begin{align}
\! \log p(\bm{y}_{1:T}) 
&\! \geq \! \mathbb{E}_{q(\bm{x}_{1:T}, \bm{f}_{2:T}, \bm{z})} \! 
\left[ \log \frac
{p(\bm{y}_{1:T}, \! \bm{x}_{1:T},\! \bm{f}_{2:T},\! \bm{z})}
{q(\bm{x}_{1:T}, \bm{f}_{2:T}, \bm{z})} \right] \nonumber \\
&=: \mathcal{L}_\text{PR-SSM}\,.
\label{eq:ELBO}
\end{align}
Maximizing the ELBO is equivalent to minimizing $\KL*{q(\bm{x}_{1:T}, \bm{f}_{2:T}, \bm{z})}{p(\bm{x}_{1:T}, \bm{f}_{2:T}, \bm{z} \mid \bm{y}_{1:T})}$ \cite{blei2017variational}, therefore this is a way to optimize the approximated model parameter distribution with respect to the intractable, true model parameter posterior.

Based on \eqref{eq:JointDistribution} and \eqref{eq:VariationalDistribution} and using standard variational calculus, the ELBO \eqref{eq:ELBO} can be transformed into
\begin{align}
\mathcal{L}_\text{PR-SSM} 
= &\sum_{t=1}^T \mathbb{E}_{q(\bm{x}_t)}[\log p(\bm{y}_t \mid \bm{x}_t)] \nonumber \\
&- \sum_{d=1}^{D_x} \KL*{q(\bm{z}_d)}{p(\bm{z}_d; \bm{\zeta}_d)}\,.
\label{eq:ELBOdetail}
\end{align}
The first part is the expected log-likelihood of the observed system outputs $\bm{y}$ based on the observation model and the variational latent state distribution $q(\bm{x})$.
This term captures the capability of the learned latent state model to explain the observed system behavior.
The second term is a regularizer on the inducing output distribution that penalizes deviations from the GP prior.
Due to this term, PR-SSM automatically trades off data fit against model complexity.
A detailed derivation of the ELBO can be found in the supplementary material.

\subsection{Stochastic Gradient ELBO Optimization}
\label{sec:StochasticELBOEvaluation}

Training the proposed PR-SSM requires maximizing the ELBO in \eqref{eq:ELBOdetail} with respect to the model parameters $\theta_\text{PR-SSM}$.
While the second term, as KL between two Gaussian distributions, can be easily computed, the first term requires evaluation of an expectation with respect to the latent state distribution $q(\bm{x})$.
Since the true non-linear, latent dynamics is maintained in the variational approximation \eqref{eq:VariationalDistribution}, analytic evaluation of $q(\bm{x})$ is still intractable.
To make this process tractable, the Markovian structure of the unobserved states and the sparse GP approximation can be exploited to enable a differentiable, sampling-based estimation of the expectation term.
Specifically, the marginal latent state distribution $q(\bm{x}_t)$ at time $t$ is \textit{conditionally} independent of past time steps, given the previous state distribution $q(\bm{x}_{t-1})$ and the explicit representation of GP inducing points.
Samples $\tilde{\bm{x}}_t \sim q(\bm{x}_t)$ can therefore be obtained by recursively drawing from the sparse GP posterior in \eqref{eq:GPPosteriorDetail} for $t = 1, \ldots, T$.
Drawing samples from a Gaussian distribution can be made differentiable with respect to its parameters $\mu_d$, $\sigma^2_d$ using the \textit{re-parametrisation trick} \cite{kingma2013auto} by first drawing samples $\epsilon \sim \mathcal{N}(0,1)$ and then computing
\begin{equation}
\tilde{x}_{t+1,d} = \mu_d(\hat{\bm{x}}_t) + \epsilon \sqrt{\sigma^2_d(\hat{\bm{x}}_t, \hat{\bm{x}}_t) + \sigma^2_{x,d}}\,,
\label{eq:Sampling}
\end{equation}
with $\hat{\bm{x}}_t = (\tilde{\bm{x}}_t, \bm{u}_t)$ and $\tilde{\bm{x}}_1 \sim q(\bm{x}_1)$.
The gradient is propagated back through time due to this re-paramatrization and unrolling of the latent state.
Using \eqref{eq:Sampling}, an unbiased estimator of the first term in the ELBO in \eqref{eq:ELBOdetail} is given by
\begin{equation}
\mathbb{E}_{q(\bm{x}_t)}[\log (\bm{y}_t \mid \bm{x}_t)] \approx \frac{1}{N}  \sum_{i=1}^N \log p(\bm{y}_t \mid \tilde{\bm{x}}_t^{(i)})\,.
\label{eq:SampleExpectationEstimator}
\end{equation}
Based on the stochastic ELBO evaluation, analytic gradients of \eqref{eq:ELBOdetail} can be derived to facilitate stochastic gradient-descent-based model optimization.

\subsection{Model Predictions}
\label{sec:ModelPredictions}

After model optimization based on the ELBO \eqref{eq:ELBOdetail}, model predictions for a new input sequence $\bm{u}_{1:T}$ and initial latent state $\bm{x}_1$ can be obtained based on the approximate, variational posterior distribution in \eqref{eq:VariationalDistribution}.
In contrast to \cite{mattos2015recurrent}, no approximations such as moment matching are required for model predictions.
Instead, the complex latent state distribution is approximated based on samples as in \eqref{eq:Sampling}.
The predicted observation distribution can then be computed from the latent distribution according to the observation model in \eqref{eq:ObservationModel}.
Instead of a fixed, uninformative initial latent state, a learned recognition model can be utilized to find a more informative model initialization (cf.\,\ref{sec:ExtensionsforLargeDatasets}).

\section{Extensions for Large Datasets}
\label{sec:ExtensionsforLargeDatasets}

\begin{figure}[t]
	%\vskip 0.2in
	\begin{center}
		\centerline{\includegraphics[width=\columnwidth]{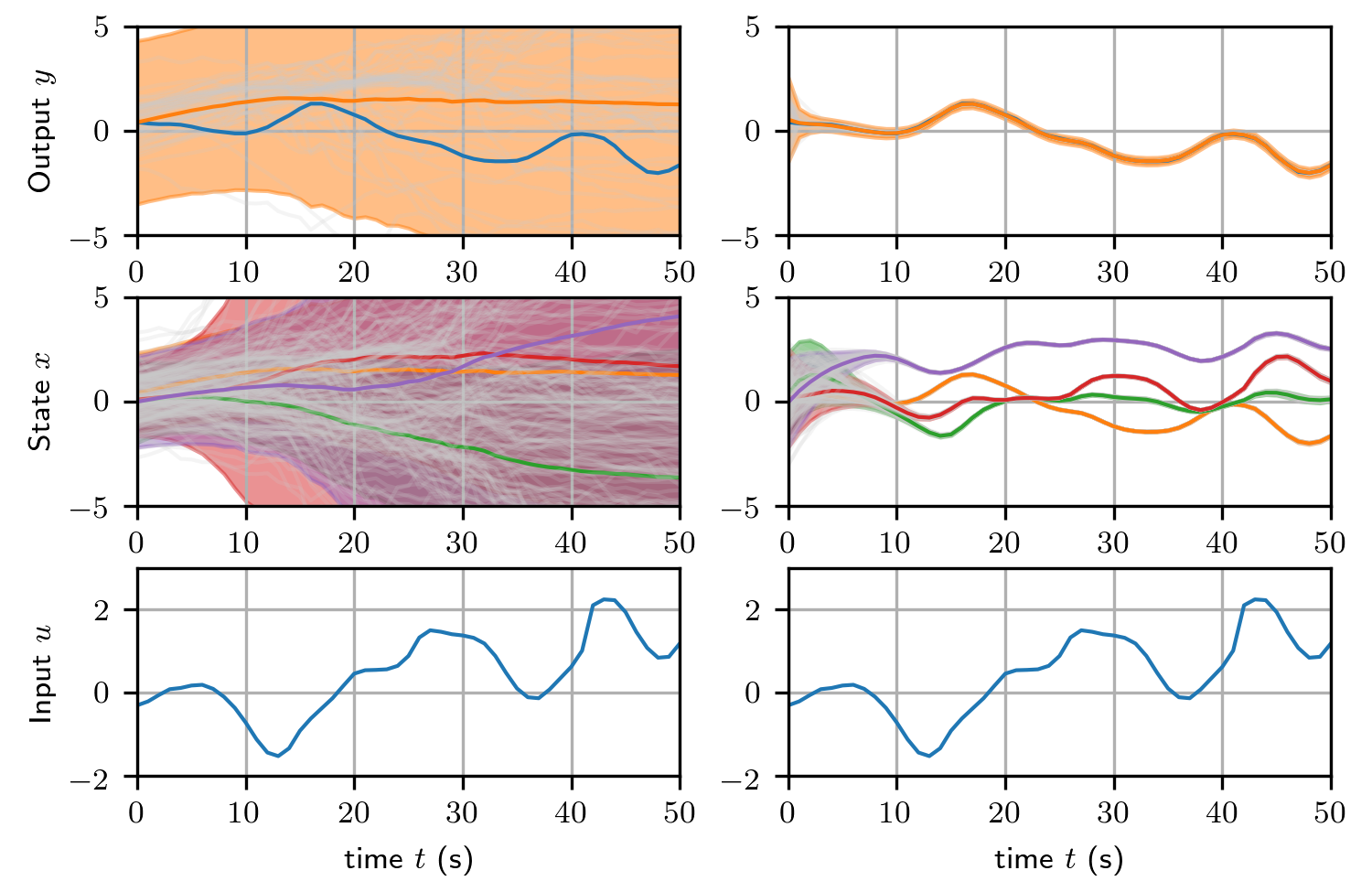}}
		\vspace{-3mm}
		\caption{Predictions of the initial, untrained (left) and the final, trained PR-SSM (right) based on the full gradient ELBO optimization.
		The system input/output data (blue) is visualized together with the model prediction (orange) for a part of the \textit{Furnace} dataset.
		Samples of the latent space distribution and output distribution are shown in gray.
		The shaded areas visualize mean +/- two std.
		The initial model exhibits a random walk behavior in the latent space.
		In the trained model, the decay of the initial state uncertainty can be observed in the first time steps.
		In this experiment, no recognition model has been used (cf.\,Sec.\,\ref{sec:ExtensionsforLargeDatasets}).}
		\label{fig:ModelComparisonUntrainedTrained}
	\end{center}
	\vskip -0.4in
\end{figure}

\begin{figure}[t]
	%\vskip 0.2in
	\begin{center}
		\centerline{\includegraphics[width=\columnwidth]{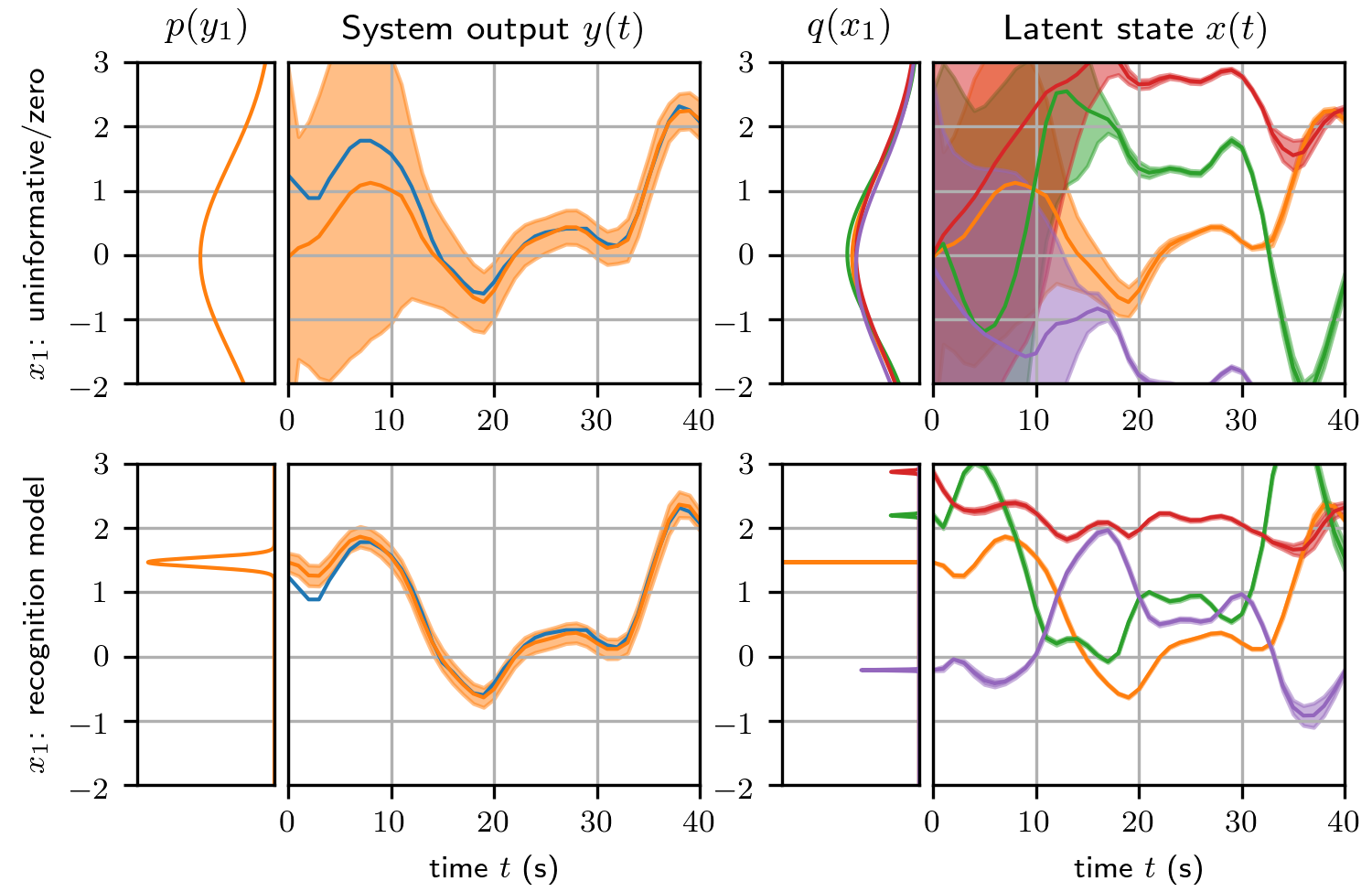}}
		\vspace{-3mm}
		\caption[Latent state initialization.]{Comparison of the fully trained PR-SSM predictions with (lower row) and without (upper row) initial state recognition model on a test dataset.
		The initial transient based on the uncertainty from an uninformative initial state distribution $q(\bm{x}_1) = \mathcal{N}(\bm{x}_1 \mid \bm{0}, \bm{I})$ decays, as shown in upper plots.
		Below the outcome is shown when $q(\bm{x}_1)$ is initialized by the smoothing distribution $q(\bm{x}_1 \mid \bm{y}_{1:L}, \bm{u}_{1:L})$, given the first $L$ steps of system input/output.
		Using the recognition model yields a significantly improved latent state initialization and therefore decreases the initial state uncertainty and the initial transient behavior.}
		\label{fig:InitialStateComparison}
	\end{center}
	\vskip -0.4in
\end{figure}

Optimizing the ELBO \eqref{eq:ELBOdetail} based on the full gradient is prohibitive for large datasets and long trajectories.
Instead, a stochastic optimization scheme based on mini-batches of sub-trajectories is introduced.

Directly optimizing the initial latent state distribution $q(\bm{x}_1)$ for each sub-trajectory would lead to a full parametrization of the latent state which is undesirable as described in Sec.\,\ref{sec:VariationalApproximation}.
Instead, we propose a parametric recognition model, which initializes the latent state $q(\bm{x}_1)$.
In recent work on SSMs \cite{mattos2015recurrent, eleftheriadis2017identification}, a recognition model is introduced to parametrize the smoothing distribution $p(\bm{x}_{1:T} \mid \bm{y}_{1:T}, \bm{u}_{1:T})$.
In this work, a similar idea is employed but only to model the initial latent state
\begin{align}
q(\bm{x}_1) = \mathcal{N}(\bm{x}_1 \mid \bm{\mu}_1, \bm{\Sigma}_1) &\approx q(\bm{x}_1 \mid \bm{y}_{1:L}, \bm{u}_{1:L})\,, \\
\bm{\mu}_1, \bm{\Sigma}_1 &= h(\bm{y}_{1:L}, \bm{u}_{1:L} ; \theta_\text{recog})\,.
\label{eq:InitialStateRecognitionModel}
\end{align}
The initial latent state distribution is approximated by a Gaussian, where mean and variance are modeled by a recognition model $h$.
The recognition model acts as a smoother, operating on the first $L$ elements of the system input/output data to infer the first latent state.
Instead of directly optimizing $q(\bm{x}_1)$ during training, errors are propagated back into the recognition model $h$, which is parametrized by $\theta_\text{recog}$.

Additionally, while $q(\bm{x}_1)$ can be inferred during training, no information about the initial latent state is available during prediction.
Thus, the recognition model is also required to enable predictions on test sequences, where the initial latent state is unknown.
\section{Experimental Evaluation}
\label{sec:ExperimentalEvaluation}

\begin{table*}[t]
	\centering
	\setlength{\tabcolsep}{1mm}
	\renewcommand{\arraystretch}{1.2}
	\caption{Comparison of model learning methods on five real-world benchmark examples.
The RMSE result (mean (std) over 5 independently learned models) is given for the free simulation on the test dataset.
For each dataset, the best result (solid underline) and second best result (dashed underline) is indicated.
The proposed PR-SSM consistently outperforms the reference (SS-GP-SSM) in the class of Markovian state space models and robustly achieves performance comparable to the one of state-of-the-art latent, autoregressive models.
Best viewed in color.}
%\vspace{-3mm}
	\label{tab:ResultsSystemIdentification}
	\vskip 0.15in
	\begin{small}
	\begin{sc}
	\begin{tabular}{lx{2cm}x{2cm}x{2cm}x{2cm}x{2cm}x{2cm}x{2cm}}
		\toprule
		 & \multicolumn{2}{x{4cm}}{One-step-ahead, autoregressive} & \multicolumn{3}{x{6cm}}{Multi-step-ahead, latent space autoregressive} & \multicolumn{2}{x{4cm}}{Markovian state-space models} \\
		\cmidrule(lr){2-3}
		\cmidrule(lr){4-6}
		\cmidrule(lr){7-8}
		Task & GP-NARX & NIGP & REVARB 1 &  REVARB 2 &  MSGP & SS-GP-SSM & \textbf{PR-SSM} \\
		\midrule
		Actuator & 
		\cellcolor[rgb]{0.988, 0.937, 0.949} 0.627 (0.005) & 
		\cellcolor[rgb]{0.937, 0.969, 0.957} 0.599 (0) & 
		\cellcolor[rgb]{0.388, 0.745, 0.482} \uline{0.438 (0.049)} & 
		\cellcolor[rgb]{0.988, 0.988, 1.000} 0.613 (0.190) & 
		\cellcolor[rgb]{0.973, 0.412, 0.420} 0.771 (0.098) & 
		\cellcolor[rgb]{0.980, 0.686, 0.698} 0.696 (0.034) & 
		\cellcolor[rgb]{0.604, 0.831, 0.671} \dashuline{0.502 (0.031)} \\
		Ballbeam &
		\cellcolor[rgb]{0.988, 0.988, 1.000} 0.284 (0.222) &
		\cellcolor[rgb]{0.514, 0.796, 0.588} \dashuline{0.087 (0)} &
		\cellcolor[rgb]{0.988, 0.988, 1.000} 0.139 (0.007) &
		\cellcolor[rgb]{0.988, 0.988, 1.000} 0.209 (0.012) &
		\cellcolor[rgb]{0.851, 0.929, 0.878} 0.124 (0.034) &
		\cellcolor[rgb]{0.973, 0.412, 0.420} 411.6 (273.0) &
		\cellcolor[rgb]{0.388, 0.745, 0.482} \uline{0.073 (0.007)}\\
		Drives &
		\cellcolor[rgb]{0.988, 0.988, 1.000} 0.701 (0.015) &
		\cellcolor[rgb]{0.388, 0.745, 0.482} \uline{0.373 (0)} &
		\cellcolor[rgb]{0.976, 0.553, 0.561} 0.828 (0.025) &
		\cellcolor[rgb]{0.973, 0.412, 0.420} 0.868 (0.113) &
		\cellcolor[rgb]{0.529, 0.800, 0.604} \dashuline{0.451 (0.021)} &
		\cellcolor[rgb]{0.988, 0.933, 0.941} 0.718 (0.009) &
		\cellcolor[rgb]{0.604, 0.831, 0.667} 0.492 (0.038)\\
		Furnace &
		\cellcolor[rgb]{0.847, 0.929, 0.875} 1.201 (0.000) &
		\cellcolor[rgb]{0.988, 0.988, 1.000} 1.205 (0) &
		\cellcolor[rgb]{0.635, 0.843, 0.694} \dashuline{1.195 (0.002)} &
		\cellcolor[rgb]{0.388, 0.745, 0.482} \uline{1.188 (0.001)} &
		\cellcolor[rgb]{0.980, 0.624, 0.631} 1.277 (0.127) &
		\cellcolor[rgb]{0.973, 0.412, 0.420} 1.318 (0.027) &
		\cellcolor[rgb]{0.984, 0.765, 0.776} 1.249 (0.029)\\
		Dryer &
		\cellcolor[rgb]{0.988, 0.949, 0.961} 0.310 (0.044) &
		\cellcolor[rgb]{0.988, 0.988, 1.000} 0.268 (0) &
		\cellcolor[rgb]{0.973, 0.412, 0.420} 0.851 (0.011) &
		\cellcolor[rgb]{0.988, 0.906, 0.914} 0.355 (0.027) &
		\cellcolor[rgb]{0.416, 0.753, 0.506} \dashuline{0.146 (0.004)} &
		\cellcolor[rgb]{0.443, 0.765, 0.529} 0.152 (0.006) &
		\cellcolor[rgb]{0.388, 0.745, 0.482} \uline{0.140 (0.018)}\\
		\midrule\
		Sarcos &
		\cellcolor[rgb]{0.973, 0.412, 0.420} 0.169 (-) & n.a. & n.a. & n.a. & n.a. & n.a. & \cellcolor[rgb]{0.388, 0.745, 0.482} \uline{0.049 (-)} \\
		\bottomrule
		\end{tabular}
	\end{sc}
	\end{small}
	\vskip -0.15in
\end{table*}

In the following, we present insights into the PR-SSM optimization schemes, a comparison to state-of-the-art model learning methods on real world datasets and results from a large scale experiment.

\subsection{PR-SSM Learning}
\label{sec:PR-SSMLearning}

For small datasets (i.e.\,short training trajectory lengths), the model can be trained based on the full gradient of the ELBO in \eqref{eq:ELBOdetail}.
A comparison of the model predictions before and after training with the full ELBO gradient is visualized in Fig.\,\ref{fig:ModelComparisonUntrainedTrained}.

Empirically, three major shortcomings of the full gradient-based optimization schemes are observed:
(i) Computing the full gradient for long trajectories is expensive and prone to the well-known problems of exploding and vanishing gradients \cite{pascanu2013difficulty}.
(ii) An uninformative initial state is prohibitive for unstable systems or systems with slowly decaying initial state transients.
(iii) Momentum-based optimizers (e.g.\,Adam) exhibit fragile optimization performance and are prone to overfitting.

The proposed method addresses these problems by employing the stochastic ELBO gradient based on minibatches of sub-trajectories and the initial state recognition model (cf. Sec.\,\ref{sec:ExtensionsforLargeDatasets}).
Fig.\,\ref{fig:InitialStateComparison} visualizes the initial state distribution $q(\bm{x}_1)$ and the corresponding predictive output distribution $p(\bm{y}_1)$ for the fully trained model based on the full gradient (top row), as well as for the model based on the stochastic gradient and recognition model (bottom row).
The transient dynamics and the associated model uncertainty is clearly visible for the first 15 time steps until the initial transient decays and approaches the true system behavior.
In contrast, the learned recognition model almost perfectly initializes the latent state, leading to much smaller deviations in the predicted observations and far less predictive uncertainty.
Notice how the recognition model is most certain about the distribution of the first latent state dimension (orange), which is directly coupled to the observation through the parametric observation model (cf.\,\eqref{eq:ObservationModel}).
The uncertainty for the remaining, latent states, in contrast, is slightly higher.

Comparing the full ELBO gradient-based model learning and the stochastic version with the recognition model, the stochastic model learning is far more robust and counteracts the overfitting tendencies in the full gradient-based model learning.
A comparison of the model learning progress for both methods is depicted in the supplementary material.
Due to the improved optimization robustness and the applicability to larger datasets, the stochastic, recognition-model-based optimization scheme is employed for the model learning benchmark presented in the next section.

\subsection{Model Learning Benchmark}
\label{sec:ModelLearningBenchmark}

\begin{figure*}[t]
	%\vskip 0.2in
	\begin{center}
		\centerline{\includegraphics[width=\textwidth]{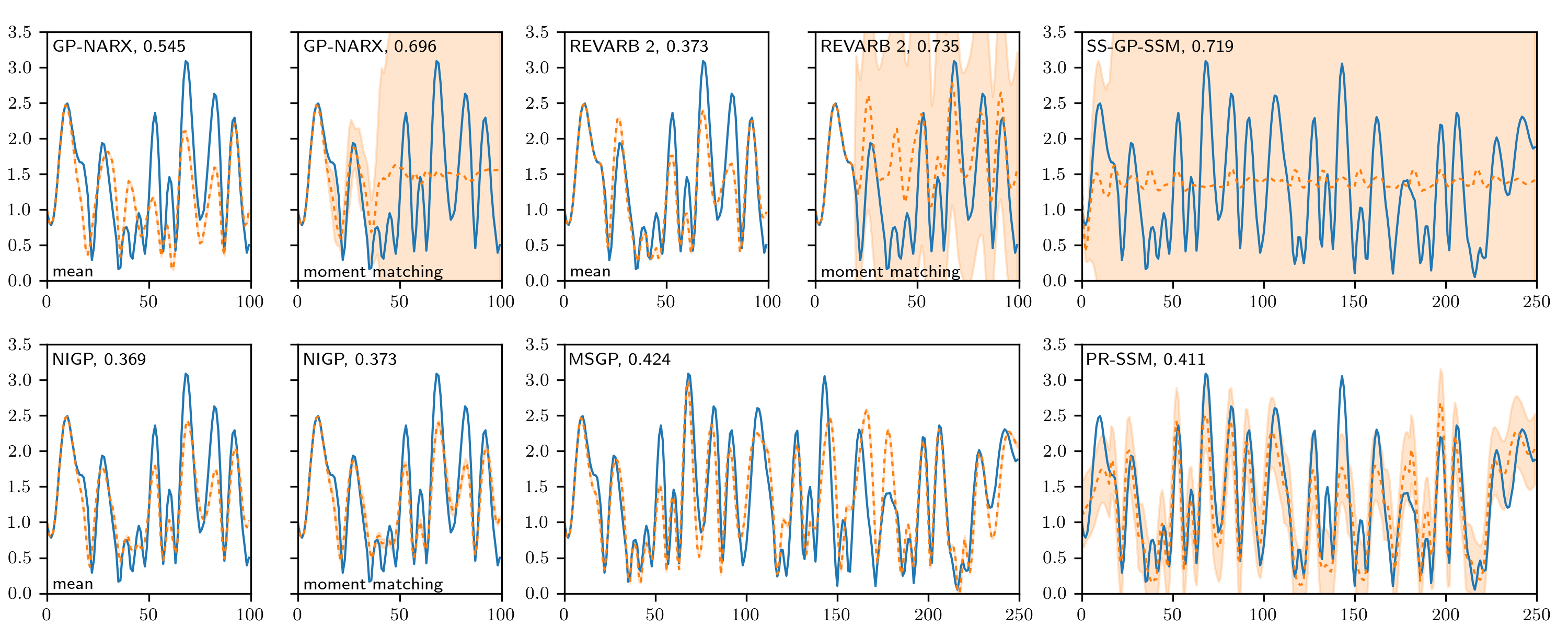}}
		\vspace{-3mm}
		\caption{Free simulation results for the benchmark methods on the \textit{Drives} test dataset. The true, observed system output (blue) is compared to the individual model's predictive output distribution (orange, mean +/- two std).
Results are presented for the one-step-ahead models GP-NARX and NIGP in the left column.
REVARB and MSGP (shown in the middle column) are both based on multi-step optimized autoregressive GP models in latent space.
In the right column, the SS-GP-SSMs, as a model based on a Markovian latent state, is compared to the proposed PR-SSM.}
		\label{fig:BenchmarkPredictionDrives}
	\end{center}
	\vskip -0.3in
\end{figure*}

The performance of PR-SSM is assessed in comparison to state-of-the-art model learning methods on several real-world datasets as previously utilized by \cite{mattos2015recurrent}.
The suite of reference methods is composed of: One-step ahead autoregressive GP models: GP-FITC \cite{snelson2006sparse} and NIGP \cite{mchutchon2011gaussian}.
Multi-step-ahead autoregressive and recurrent GP models in latent space: REVARB based on 1 respectively 2 hidden layers \cite{mattos2015recurrent} and MSGP \cite{doerr2017optimizing}.
GP-SSMs, based on a full Markovian state: SS-GP-SSM \cite{svensson2017flexible} and the proposed PR-SSM.
Currently, no published and runnable code exists for the model learning frameworks presented in \cite{turner2010state, frigola2013bayesian, frigola2014variational, eleftheriadis2017identification}.
Reproducing these methods is not straightforward and outside the scope of this work.
To enable a fair comparison, all methods is given access to a predefined amount of input/output data for initialization.
Details about the benchmark methods, their configuration, as well as the benchmark datasets can be found in the supplementary material.

The benchmark results are summarized in Tab.\,\ref{tab:ResultsSystemIdentification}.
A detailed visualization of the resulting model predictions on the \textit{Drives} dataset is shown in Fig.\,\ref{fig:BenchmarkPredictionDrives}.
For the one-step-ahead models (GP-NARX, NIGP), two variants are used to obtain long-term predictive distributions:
Propagating the mean (no uncertainty propagation) and approximating the true posterior by a Gaussian using exact moment matching \cite{girard2003multiple}.
The results show that PR-SSM consistently outperforms the SS-GP-SSM learning method.
Similarly, performance is improved in comparison to baseline methods (GP-NARX and NIGP).
In the ensemble of models based on long-term optimized autoregressive structure (REVARB, MSGP), no method is clearly superior.
However, the performance of PR-SSM is in all cases close to the one of the best performing method.
Note that PR-SSM demonstrates robust model learning performance and does not exhibit severe failure any dataset as some of the contestants do.

\subsection{Large Scale Experiment}
\label{sec:LargeScaleExperiment}

\begin{figure}[t]
	%\vskip 0.2in
	\begin{center}
		\centerline{\includegraphics[width=\columnwidth]{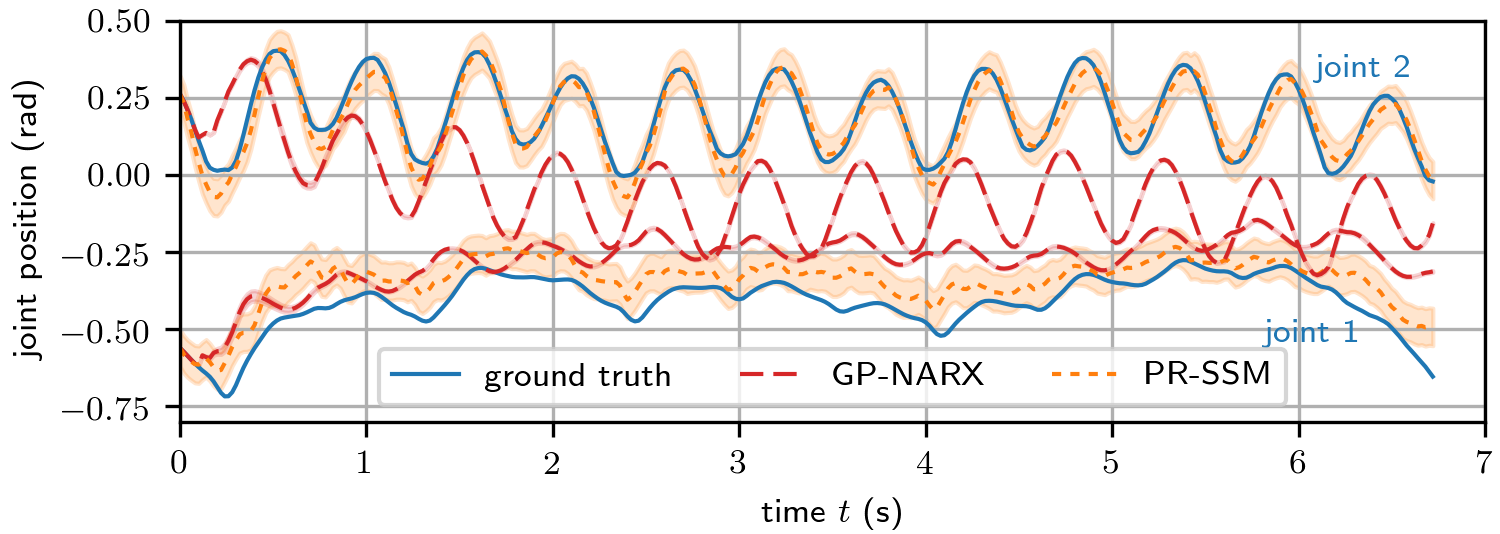}}
		\vspace{-3mm}
		\caption{Results on the Sarcos large scale task: Predictions from the GP-NARX baseline (red) and the PR-SSM (orange) for two out of seven joint positions.
		The ground truth, measured joint positions are shown in blue.
		PR-SSM clearly improves over the GP-NARX predictions.
		Similar results are obtained for PR-SSM on the remaining 5 joints, where the GP-NARX model fails completely (cf.\,supplementary materials for details).}
		\label{fig:LargeScale}
	\end{center}
	\vskip -0.4in
\end{figure}

To evaluate the scalability, results are provided for the forward dynamics model of the SARCOS 7 degree of freedom robotic arm.
The task is characterized by 60 experiments of length 337 (approx 20.000 datapoints), 7 input, and 7 output dimensions.
PR-SSM is set up with a latent state dimensionality $D_x = 14$.
From the set of reference methods, only the GP-NARX model can be adapted to run efficiently on this dataset without major efforts.
The details about the task and method configuration are given in the supplementary material.
A visualization of the model predictions is shown in Fig\,\ref{fig:LargeScale} and prediction RMSEs are listed in Tab.\,\ref{tab:ResultsSystemIdentification}.
The results show that PR-SSM is able to robustly and accurately learn the forward dynamics for all system outputs from all experimental data.
In contrast, the GP-NARX baseline achieves worse predictions and fails to predict the remaining five joints (not shown).

\section{Conclusion}
\label{sec:Conclusion}

In this work, we presented Probabilistic Recurrent State-Space Models (PR-SSM), a novel model structure and efficient inference scheme for learning probabilistic, Markovian state-space models.
Based on GP priors and doubly stochastic variational inference, a novel model optimization criterion is derived, which is closely related to the one of powerful, but deterministic, RNNs or LSTMs.
By maintaining the true latent state distribution and thereby enabling long-term gradients, efficient inference in latent spaces becomes feasible.
Furthermore, a novel recognition model enables learning of unstable or slow dynamics as well as scalability to large datasets.
Robustness, scalability and high performance in model learning is demonstrated on real-world datasets in comparison to state-of-the-art methods.

A limitation of PR-SSM is its dependency on an a-priori fixed latent state dimensionality.
This shortcoming could potentially be resolved by a sparsity enforcing latent state prior, which would suppress unnecessary latent state dimensions.

\section*{Acknowledgements}
\label{sec:Acknowledgements}

This research was supported in part by National Science Foundation grants IIS-1205249, IIS-1017134, EECS-0926052, the Office of Naval Research, the Okawa Foundation, and the Max-Planck-Society.

\bibliography{references}
\bibliographystyle{icml2018}

\clearpage
\begin{appendix}
%\section{Supplementary Material}
%\label{sec:SupplementaryMaterial}

This supplementary material provides details about the derivations and configuration of the proposed PR-SSM in Sec.\,\ref{sec:ProbabilisticRecurrentStateSpaceModelSupp}.
Sec.\,\ref{sec:ModelLearningBenchmarkDetails} elaborates on the reference methods and the employed datasets in the model learning benchmark.
Finally, additional experimental results from PR-SSM learning, the model learning benchmark, and the large scale experiment are summarized in Sec.\,\ref{sec:AditionalResults}.

\section{Probabilistic Recurrent State-Space Model}
\label{sec:ProbabilisticRecurrentStateSpaceModelSupp}

\subsection{Evidence Lower Bound (ELBO)}
\label{sec:ELBO}

Summarizing the model assumptions from the main paper, the model's joint distribution is given by
\begin{align}
\label{eq:JointDistributionSupp}
p(\bm{y}_{1:T}, \bm{x}_{1:T}, \bm{f}_{2:T}, \bm{z}) = 
& \left[ \prod_{t=1}^T p(\bm{y}_t \mid \bm{x}_t) \right] \nonumber \\
& \left[ \prod_{t=2}^T p(\bm{x}_t \mid \bm{f}_t) \right] \nonumber \\
& \left[ \prod_{t=2}^T \prod_{d=1}^{D_x} p(f_{t,d} \mid \hat{\bm{x}}_{t-1}, \bm{z}_d) p(\bm{z}_d) \right] \nonumber \\
& p(\bm{x}_1)\,.
\end{align}
The variational distribution over the unknown model variables is defined as
\begin{align}
\label{eq:VariationalDistributionSupp}
q(\bm{x}_{1:T}, \bm{f}_{2:T}, \bm{z}) 
= 
& \left[ \prod_{t=2}^T p(\bm{x}_t \mid \bm{f}_t) \right] \nonumber \\
& \left[ \prod_{t=2}^T \prod_{d=1}^{D_x} p(f_{t, d} \mid \hat{\bm{x}}_{t-1}, \bm{z}_d) q(\bm{z}_d) \right] \nonumber \\
& q(\bm{x}_1)\,.
\end{align}
Together, the derivation of the ELBO is given below in \eqref{eq:ELBOSupp1} to \eqref{eq:ELBOSupp}.
\begin{figure*}
\begin{align}
\! \log p(\bm{y}_{1:T}) 
&\! \geq \! \mathbb{E}_{q(\bm{x}_{1:T}, \bm{f}_{2:T}, \bm{z})} \! 
\left[ \log \frac
{p(\bm{y}_{1:T}, \! \bm{x}_{1:T},\! \bm{f}_{2:T},\! \bm{z})}
{q(\bm{x}_{1:T}, \bm{f}_{2:T}, \bm{z})} \right] \label{eq:ELBOSupp1} \\
& = 
\mathbb{E}_{q(\bm{x}_{1:T}, \bm{f}_{2:T}, \bm{z})}
\left[
\log \frac
{\left[ \prod_{t=1}^T p(\bm{y}_t \mid \bm{x}_t) \right] \left[ \prod_{t=2}^T p(\bm{x}_t \mid \bm{f}_t) \right] \left[ \prod_{t=2}^T \prod_{d=1}^{D_x} p(f_{t,d} \mid \hat{\bm{x}}_{t-1}, \bm{z}_d) p(\bm{z}_d) \right] p(\bm{x}_1)}
{\left[ \prod_{t=2}^T p(\bm{x}_t \mid \bm{f}_t) \right] \left[ \prod_{t=2}^T \prod_{d=1}^{D_x} p(f_{t, d} \mid \hat{\bm{x}}_{t-1}, \bm{z}_d) q(\bm{z}_d) \right] q(\bm{x}_1)}
\right]
\\
& = 
\mathbb{E}_{q(\bm{x}_{1:T}, \bm{f}_{2:T}, \bm{z})}
\left[
\log \frac
{\left[ \prod_{t=1}^T p(\bm{y}_t \mid \bm{x}_t) \right] \left[\prod_{d=1}^{D_x} p(\bm{z}_d) \right] p(\bm{x}_1)}
{\left[ \prod_{d=1}^{D_x} q(\bm{z}_d) \right] q(\bm{x}_1)}
\right]
\\
& = 
\mathbb{E}_{q(\bm{x}_{1:T}, \bm{f}_{2:T}, \bm{z})}
\left[
\log 
\prod_{t=1}^T p(\bm{y}_t \mid \bm{x}_t)
\right]
+
\mathbb{E}_{q(\bm{x}_{1:T}, \bm{f}_{2:T}, \bm{z})}
\left[
\sum_{d=1}^{D_x} \log \frac
{p(\bm{z}_d)}
{q(\bm{z}_d)}
\right]
+
\mathbb{E}_{q(\bm{x}_{1:T}, \bm{f}_{2:T}, \bm{z})}
\left[
\log \frac
{p(\bm{x}_1)}
{q(\bm{x}_1)}
\right]
\\
& = 
\mathbb{E}_{q(\bm{x}_{1:T})}
\left[
\log 
\prod_{t=1}^T p(\bm{y}_t \mid \bm{x}_t)
\right]
+
\mathbb{E}_{q(\bm{z})}
\left[
\sum_{d=1}^{D_x}\log \frac
{p(\bm{z}_d)}
{q(\bm{z}_d)}
\right]
+
\mathbb{E}_{q(\bm{x}_{1})}
\left[
\log \frac
{p(\bm{x}_1)}
{q(\bm{x}_1)}
\right]
\\
& = 
\sum_{t=1}^T
\mathbb{E}_{q(\bm{x}_{t})}
\left[
\log p(\bm{y}_t \mid \bm{x}_t)
\right]
+
\sum_{d=1}^{D_x} \KL*{q(\bm{z}_d)}{p(\bm{z}_d)}
+
\KL*{q(\bm{x}_1)}{p(\bm{x}_1)}
\label{eq:ELBOSupp}
\end{align}
\end{figure*}

In the ELBO, as derived in \eqref{eq:ELBOSupp}, the last term is a regularization on the initial state distribution.
For the full gradient-based optimization in the main paper, an uninformative initial distribution is chosen and fixed, such that the third term is dropped.
In the stochastic optimization scheme, this term acts as a regularization preventing the recognition model to become overconfident in its predictions.

\subsection{Model Configuration}
\label{sec:ModelConfiguration}

The PR-SSM exhibits a large number of model (hyper-) parameters $\theta_\text{PR-SSM}$ which need to be initialized.
However, empirically, most of these model parameters can be initialized to a default setting as given in Tab.\,\ref{tab:ModelParameterConfiguration}.
This default configuration has been employed for all benchmark experiments presented in the main paper.

\begin{table}[t]
	\caption{Default configuration for the initialization of the PR-SSM (hyper-) parameters $\theta_\text{PR-SSM}$.
	This configuration has been employed for all experiments in the benchmark section.}
	\label{tab:ModelParameterConfiguration}
	\vskip 0.15in
	\begin{center}
	\begin{small}
	\begin{sc}
  \begin{tabular}{p{2.5cm}l}
		\toprule
		Parameter & Initialization \\
		\midrule
		Inducing inputs & $\bm{\zeta}_d \sim \mathcal{U}(-2, 2) \in \mathbb{R}^{P \times (D_x + D_u)}$ \\
		\midrule
		\multirow{3}{2.5cm}{Inducing outputs} & $q(\bm{z}_d) = \mathcal{N}(\bm{z}_d \mid \bm{\mu}_d, \bm{\Sigma}_d) \in \mathbb{R}^P$ \\
		 & $\mu_{d,i} \sim \mathcal{N}(\mu_{d,i} \mid 0, 0.05^2)$ \\
		%\quad \forall i \in [1,N_\text{inducing}]$ \\
		 & $\bm{\Sigma}_d = 0.01^2 \cdot \bm{I}$ \\
		\midrule
		Process noise & $\sigma_{\text{x},i}^2 = 0.002^2 \quad \forall i \in [1,D_\text{x}]$ \\
		Sensor noise & $\sigma_{\text{y},i}^2 = 1.0^2 \quad \forall i \in [1,D_\text{y}]$ \\
		\midrule
		\multirow{2}{2.5cm}{kernel hyper-parameters} & $\sigma_f^2 = 0.5^2$ \\
		& $l_i^2 = 2 \quad \forall i \in [1, D_x]$ \\
		\bottomrule
		\end{tabular}
	\end{sc}
	\end{small}
	\end{center}
	\vskip -0.1in
\end{table}

\begin{table}[t]
	\caption{Structural configuration of the PR-SSM as utilized in the benchmark experiments.}
	\label{tab:ModelStructureConfiguration}
	\vskip 0.15in
	\begin{center}
	\begin{small}
	\begin{sc}
	\begin{tabular}{p{2.5cm}l}
		\toprule
		Parameter & Initialization \\
		\midrule
		Inducing points & $P = 20$ \\
		State samples & $N = 50$ \\
		\midrule
		\multirow{2}{2.5cm}{Subtrajectories} & $N_\text{batch} = 10$ \\
		& $T_{sub} = 100$ \\
		\midrule
		Latent space & $D_x = 4$ \\
		\bottomrule
		\end{tabular}
	\end{sc}
	\end{small}
	\end{center}
	\vskip -0.1in
\end{table}

The PR-SSM's latent state dynamics model and noise models are configured to initially exhibit a random walk behavior.
This behavior is clearly visible for the prediction based on the untrained model in Fig.\,2 of the main paper.
The GP prior is approximating the identity function based on an identity mean function and almost zero inducing outputs (up to a small Gaussian noise term to avoid singularities).
The inducing inputs are spread uniformly over the function's domain.
The noise processes are initializes such as to achieve high correlations between latent states over time (i.e.\,small process noise magnitude).
At the same time, a larger observation noise is required to obtain a inflation of predictive uncertainty over time.
This inflation of predictive uncertainty is again clearly visible in Fig.\,2 of the main paper.
Both noise terms are chosen in a way to obtain numerically stable gradients for both the sample based log likelihood and the backpropagation through time in the ELBO evaluation.

The number of samples used in the ELBO approximation, number of inducing points in the GP approximation and batch size are, in contrast, a trade-off between model accuracy and computational speed.
The proposed default configuration empirically showed good performance whilst being computationally tractable.

Two tuning parameters remain, which are problem specific and have to be chosen for each dataset individually.
Depending on the true system's timescales/sampling frequency and system order, the length of subtrajectories $T_{sub}$ for minibatching and the latent state dimensionality $D_x$ have to be specified manually.
For the benchmark datasets we choose $T_{sub} = 100$ and $D_x = 4$.

\section{Model Learning Benchmark Details}
\label{sec:ModelLearningBenchmarkDetails}

In the main paper, the proposed PR-SSM's long-term predictive performance is compared to several state-of-the-art methods.
The benchmark is set up similar to the evaluation presented in \cite{doerr2017optimizing}.
Details about the individual benchmark methods, their configuration and the employed datasets can be found in the following sections.
Minor adjustments with respect to the set up in \cite{doerr2017optimizing} will be pointed out in the following.
These have been introduced to enable fair comparison between all benchmark methods.

\begin{table}[t]
	\caption{Summary of the real-world, non-linear system identification benchmark tasks.
	All datasets are generated by recording input/output data of actual physical plants.
	For each dataset, the lengths of training and test set are given together with the number of past input and outputs used for the NARX dynamics models.}
	\label{tab:SysIDBenchmarkDatasets}
	\vskip 0.15in
	\begin{center}
	\begin{small}
	\begin{sc}
	\begin{tabular}{lccc}
		\toprule
		 & $N_{train}$ & $N_{test}$ & $L_u$, $L_y$ \\
		\midrule
		Actuator~\cite{dtuDataset} & 512 & 512 & 10 \\
		Ballbeam~\cite{LeuvenDataset} & 500 & 500 & 10 \\
		Drives~\cite{itDataset} & 250 & 250 & 10 \\
		Furnace~\cite{LeuvenDataset} & 148 & 148 & 3 \\
		Dryer~\cite{LeuvenDataset} & 500 & 500 & 2 \\
		\bottomrule
	\end{tabular}
	\end{sc}
	\end{small}
	\end{center}
	\vskip -0.1in
\end{table}

\subsection{Benchmark Methods}
\label{sec:BenchmarkMethods}

The proposed PR-SSM is evaluated in comparison to methods from three classes:
one-step ahead autoregressive models (GP-NARX, NIGP), multi-step ahead autoregressive models in latent space (REVARB, MSGP) and Markov state-space models (SS-GP-SSM).
To enable a fair comparison, all methods have access to a predefined amount of input/output data for initialization.

(i) GP-NARX \citep{kocijan2005dynamic}:
The system dynamics is modeled as $y_{t+1} = f(y_t, \ldots, y_{t-L_y}, u_t, \ldots, u_{t-L_u})$ with a GP prior on $f$.
The GP has a zero mean function and a squared exponential kernel with automatic relevance determination.
The kernel hyper-parameters, signal variance and lengthscales, are optimized based on the standard maximum likelihood objective.
A sparse approximation \cite{snelson2006sparse}, based on 100 inducing inputs is employed.
Moment matching \cite{girard2003multiple} is employed to obtain a long-term predictive distribution.

(ii) NIGP \citep{mchutchon2011gaussian}:
Noise Input GPs (NIGP) account for uncertainty in the input by treating input points as deterministic and inflating the corresponding output uncertainty, leading to state dependent noise, i.e. heteroscedastic GPs.
The experimental results are based on the publicly available Matlab code.
Since no sparse version is available, training is performed on the full training dataset.
Training on the full dataset is however not possible for larger datasets and provides an advantage to NIGP.
Experiments based on a random data subset of size 100 lead to decreased performance in the order of the GP-NARX results or worse.

(iii) REVARB \citep{mattos2015recurrent}:
Recurrent Variational Bayes (REVARB) is a recent proposition to optimize the lower bound to the log-marginal likelihood $\log p(\bm{y})$ using variational techniques.
This framework is based on the variational sparse GP framework~\citep{titsias2009variational}, but allows for computation of (time-)recurrent GP structures and deep GP structures (stacking multiple GP-layers in each time-step).
For our benchmark, we run REVARB using one (REVARB1) respectively two (REVARB2) hidden layers, where each layer is provided with 100 inducing inputs.
We closely follow the original setup as presented by~\citep{mattos2015recurrent}, performing 50 initial optimization steps based on fixed variances and up to 10000 steps based on variable variances.
Unlike for the other benchmark methods, the autoregressive history of REVARB implicitly becomes longer when introducing additional hidden layers.

(iv) MSGP \citep{doerr2017optimizing}:
MSGP is a GP-NARX model operating in a latent, noise free state, which is trained by optimizing its long-term predictions.
The experimental results are obtained according to the configuration described in \cite{doerr2017optimizing}, again using 100 inducing points and moment matching.

(v) SS-GP-SSM \citep{svensson2017flexible}:
The Sparse-Spectrum GP-SSM is employing a sparse spectrum GP approximation to model the system's transition dynamics in a Markovian, latent space.
The available Matlab implementation is restricted to a 2D latent space.
In the experimental results, a default configuration is employed as given by: $K = 2000, N = 40, n\_basis\_u = n\_basis\_x = 7$.
The variables are defined as given in the code published for \citep{svensson2017flexible}.

\subsection{Benchmark Datasets}
\label{sec:BenchmarkDatasets}

The benchmarks datasets are composed of popular system identification datasets from related work~\citep{narendra1990identification, kocijan2005dynamic, mattos2016latent}.
They incorporate measured input output data from technical systems like hydraulic actuators, furnaces, hair dryers or electrical motors.
For all of these problems, both inputs and outputs are one-dimensional $D_u = D_y = 1$.
However, the system's true state is higher dimensional such that an autoregressive history or an explicit latent state representation is required to capture the relevant dynamics.
The number of historic inputs and outputs for the autoregressive methods is fixed a-priori for each dataset as previously used in other publications.
For model training, datasets are normalized to zero mean and variance one based on the available training data.
References to the individual datasets, training and test trajectory length, and the utilized history for the GP-NARX models are summarized in Tab.\,\ref{tab:SysIDBenchmarkDatasets}.

\section{Additional Results}
\label{sec:AditionalResults}

\begin{figure}[t]
	\vskip 0.2in
	\begin{center}
		\centerline{\includegraphics[width=\columnwidth]{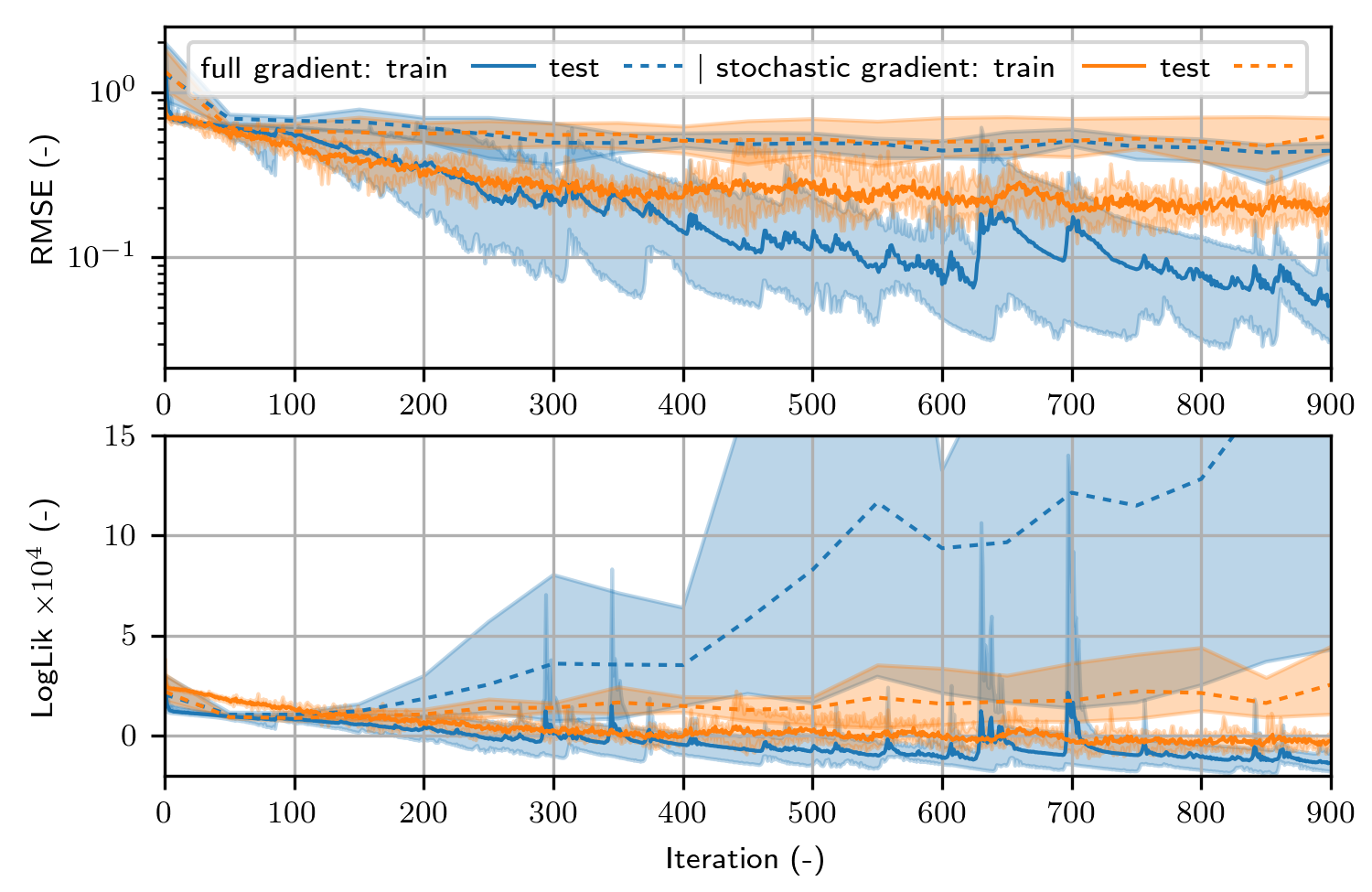}}
		\vspace{-3mm}
		\caption{Comparison of the learning progress of the proposed method on the \textit{Drive} dataset given the full ELBO gradient (blue) and the stochastic gradient, based on minibatches and the recognition model (orange).
		RMSE and log likelihood results over learning iterations are shown for the free simulation on training and test dataset.
		The full gradient optimization scheme overfitts (in particular visible in the log likelihood) and exposes a difficult optimization objective (cf.\,spikes in model loss).
		Stochastically optimizing the model-based on the proposed minibatched ELBO estimates and employing the recognition model significantly reduces overfitting and leads to more robust learning.}
		\label{fig:OverfittingComparison}
	\end{center}
	\vskip -0.4in
\end{figure}

\begin{figure*}[t]
	\vskip 0.2in
	\begin{center}
		\centerline{\includegraphics[width=\textwidth]{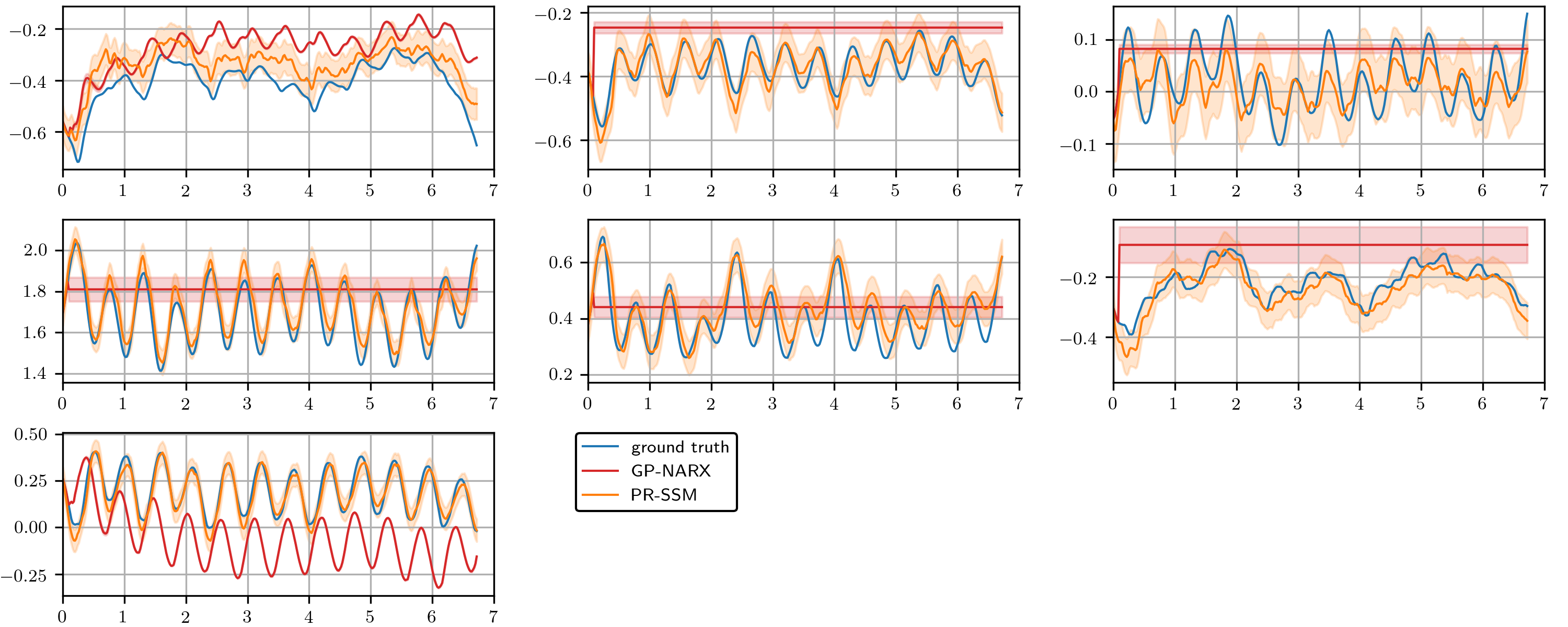}}
		\vspace{-3mm}
		\caption{Detailed results from the Sarcos large scale task: Predictions from the GP-NARX model (red) and the PR-SSM (green) for all seven joint positions as obtained for the first test experiment.
		The ground truth, measured joint positions are shown in blue.
		PR-SSM is clearly able to capture the robot arm dynamics, whereas the GP-NARX model only succesfully captures a rough model of the robot arm dynamics for two out of seven joints.}
		\label{fig:LargeScaleResultsDetails}
	\end{center}
	\vskip -0.4in
\end{figure*}

\subsection{Optimization Schemes Comparison}
\label{sec:OptimizationSchemesComparison}

In Fig.\,\ref{fig:OverfittingComparison}, the RMSE and the negative log likelihood, which is obtained for the model's long-term prediction, is depicted over learning iterations for the training- (solid line) and test- (dotted line) set from the \textit{Drives} dataset.
The full gradient optimization (blue) obtains smaller training loss in comparison to the stochastic optimization scheme for both RMSE and negative log likelihood.
The resulting test performance however indicates similar performance in terms of RMSE whilst showing clear overfitting of the full-gradient-based model in terms of log likelihood.
Additionally, optimizing, based on the full gradient, is much more delicate and less robust as indicated by the spikes in loss and the higher variance of incurred optimization progress.
Fig.\,\ref{fig:OverfittingComparison} depicts mean (lines) and minimum to maximum intervals (shaded areas) of incurred loss, based on five independent model trainings.

\subsection{Detailed Benchmark Results}
\label{sec:DetailedBenchmarkResults}

{\setlength{\tabcolsep}{1mm}
\begin{table*}[t]
\caption{Comparison of model learning methods on five real-world benchmark examples.
The RMSE result (mean (std) over 5 independently learned models) is given for the free simulation on the test dataset. For each dataset, the best result (solid underline) and second best result (dashed underline) is indicated. 
The proposed PR-SSM consistently outperforms the reference (SS-GP-SSM) in the class of Markovian state space models and robustly achieves performance comparable to the one of state-of-the-art latent, autoregressive models.}
\label{tab:ResultsSystemIdentificationSupp}
\vskip 0.15in
\begin{center}
\begin{small}
\begin{sc}
\begin{tabular}{l|x{1.9cm}x{1.4cm}|x{2cm}x{2cm}x{2cm}|x{2.8cm}x{2cm}}
		\toprule
		 & \multicolumn{2}{x{3.3cm}|}{One-step-ahead autoregressive} & \multicolumn{3}{x{6cm}|}{Multi-step-ahead autoregressive in latent space} & \multicolumn{2}{x{4.8cm}}{Markovian State-Space Models} \\
		\midrule
		 & \multicolumn{4}{l|}{Data unnormalized + Mean prediction} & \multicolumn{3}{l}{Default configuration} \\
		\midrule
		Task & GP-NARX & NIGP & REVARB 1 &  REVARB 2 &  MSGP & SS-GP-SSM & \textbf{PR-SSM} \\
		\midrule
		Actuator & 0.645 (0.018) & 0.752 (0) & \uline{0.496 (0.057)} & 0.565 (0.047) & 0.771 (0.098) & 0.696 (0.034) & \dashuline{0.502 (0.031)}\\
		Ballbeam & 0.169 (0.005) & 0.165 (0) & 0.138 (0.001) & \uline{0.073 (0.000)} & 0.124 (0.034) & 411.550 (273.043) & \dashuline{0.073 (0.007)}\\
		Drives & 0.579 (0.004) & \dashuline{0.378 (0)} & 0.718 (0.081) & \uline{0.282 (0.031)} & 0.451 (0.021) & 0.718 (0.009) & 0.492 (0.038)\\
		Furnace & \dashuline{1.199 (0.001)} & \uline{1.195 (0)} & 1.210 (0.034) & 1.945 (0.016) & 1.277 (0.127) & 1.318 (0.027) & 1.249 (0.029)\\
		Dryer & 0.278 (0.003) & 0.281 (0) & 0.149 (0.017) & \uline{0.128 (0.001)} & 0.146 (0.004) & 0.152 (0.006) & \dashuline{0.140 (0.018)}\\
		\midrule
		 & \multicolumn{4}{l|}{Data unnormalized + Moment matching} & \multicolumn{3}{l}{Default configuration} \\
		\midrule
		Task & GP-NARX & NIGP & REVARB 1 &  REVARB 2 &  MSGP & SS-GP-SSM & \textbf{PR-SSM} \\
		\midrule
		Actuator & 0.633 (0.018) & 0.601 (0) & \uline{0.430 (0.026)} & 0.618 (0.047) & 0.771 (0.098) & 0.696 (0.034) & \dashuline{0.502 (0.031)}\\
		Ballbeam & 0.077 (0.000) & 0.078 (0) & 0.131 (0.005) & \uline{0.073 (0.000)} & 0.124 (0.034) & 411.550 (273.043) & \dashuline{0.073 (0.007)}\\
		Drives & 0.688 (0.003) & \uline{0.398 (0)} & 0.801 (0.032) & 0.733 (0.087) & \dashuline{0.451 (0.021)} & 0.718 (0.009) & 0.492 (0.038)\\
		Furnace & 1.198 (0.002) & \dashuline{1.195 (0)} & \uline{1.192 (0.002)} & 1.947 (0.032) & 1.277 (0.127) & 1.318 (0.027) & 1.249 (0.029)\\
		Dryer & 0.284 (0.003) & 0.280 (0) & 0.878 (0.016) & \uline{0.123 (0.002)} & 0.146 (0.004) & 0.152 (0.006) & \dashuline{0.140 (0.018)}\\
		\midrule
		 & \multicolumn{4}{l|}{Data normalization + Mean prediction} & \multicolumn{3}{l}{Default configuration} \\
		\midrule
		Task & GP-NARX & NIGP & REVARB 1 &  REVARB 2 &  MSGP & SS-GP-SSM & \textbf{PR-SSM} \\
		\midrule
		Actuator & 0.665 (0.014) & 0.791 (0) & \dashuline{0.506 (0.092)} & 0.559 (0.069) & 0.771 (0.098) & 0.696 (0.034) & \uline{0.502 (0.031)}\\
		Ballbeam & 0.357 (0.199) & 0.154 (0) & 0.141 (0.004) & 0.206 (0.008) & \dashuline{0.124 (0.034)} & 411.550 (273.043) & \uline{0.073 (0.007)}\\
		Drives & 0.564 (0.029) & \uline{0.369 (0)} & 0.605 (0.027) & \dashuline{0.376 (0.026)} & 0.451 (0.021) & 0.718 (0.009) & 0.492 (0.038)\\
		Furnace & 1.201 (0.001) & 1.205 (0) & \dashuline{1.196 (0.002)} & \uline{1.189 (0.001)} & 1.277 (0.127) & 1.318 (0.027) & 1.249 (0.029)\\
		Dryer & 0.282 (0.001) & 0.269 (0) & \dashuline{0.123 (0.001)} & \uline{0.113 (0)} & 0.146 (0.004) & 0.152 (0.006) & 0.140 (0.018)\\
		\midrule
		 & \multicolumn{4}{l|}{Data normalization + Moment matching} & \multicolumn{3}{l}{Default configuration} \\
		\midrule
		Task & GP-NARX & NIGP & REVARB 1 &  REVARB 2 &  MSGP & SS-GP-SSM & \textbf{PR-SSM} \\
		\midrule
		Actuator & 0.627 (0.005) & 0.599 (0) & \uline{0.438 (0.049)} & 0.613 (0.190) & 0.771 (0.098) & 0.696 (0.034) & \dashuline{0.502 (0.031)}\\
		Ballbeam & 0.284 (0.222) & \dashuline{0.087 (0)} & 0.139 (0.007) & 0.209 (0.012) & 0.124 (0.034) & 411.550 (273.043) & \uline{0.073 (0.007)}\\
		Drives & 0.701 (0.015) & \uline{0.373 (0)} & 0.828 (0.025) & 0.868 (0.113) & \dashuline{0.451 (0.021)} & 0.718 (0.009) & 0.492 (0.038)\\
		Furnace & 1.201 (0.000) & 1.205 (0) & \dashuline{1.195 (0.002)} & \uline{1.188 (0.001)} & 1.277 (0.127) & 1.318 (0.027) & 1.249 (0.029)\\
		Dryer & 0.310 (0.044) & 0.268 (0) & 0.851 (0.011) & 0.355 (0.027) & \dashuline{0.146 (0.004)} & 0.152 (0.006) & \uline{0.140 (0.018)}\\
		\bottomrule
\end{tabular}
\end{sc}
\end{small}
\end{center}
\vskip -0.1in
\end{table*}
}

In Tab.\,\ref{tab:ResultsSystemIdentificationSupp}, detailed results are provided for the benchmark experiments.
The reference learning methods in the presented benchmark are highly deceptive to changes in the data pre-processing and the long-term prediction method.
Therefore, results are detailed for GP-NARX, NIGP, REVARB 1, and REVARB 2 for all combinations of normalized/unnormalized training data and mean or moment matching predictions.
The results for methods MSGP, SS-GP-SSM and PR-SSM are always computed for the normalized datasets using the method specific propagation of uncertainty schemes.

Obtaining uncertainty estimates is one key requirement for employing the long-term predictions, e.g. in model-based control.
Therefore, only the predictive results based on the approximate propagation of uncertainty through moment matching is considered in the main paper, although better results in RMSE are sometimes obtained from employing only the mean predictions.
A comparison of the predictive results based on mean and moment matching predictions on the \textit{Drives} dataset is shown in Fig.\,5.
The results from the unnormalized datasets and moment matching are in line with the results published in \cite{doerr2017optimizing}.

\subsection{Large Scale Experiment Details}
\label{sec:LargeScaleExperimentDetails}

The \textit{Sarcos} task is based on a publicly available dataset comprising joint positions, velocities, acceleration and torques of a seven degrees-of-freedom SARCOS anthropomorphic robot arm.
This dataset has been previously used in \cite{vijayakumar2000lwpr, williams2006gaussian} in the task of learning the system's inverse dynamics, therefore mapping joint position, velocities, and accelerations to the required joint torques.
This task can be framed as a standard regression problem, which is solved in a supervised fashion.
In contrast, in this paper, we consider the task of learning the forward dynamics, i.e. predicting the joint positions given a sequence of joint torques.
The system output is therefore given by the seven joint positions ($D_y = 7$).
Joint velocities and acceleration, as latent states, are not available for learning but have to be inferred.
The system input is given by the seven joint torques ($D_u = 7$).

The original training dataset (44.484 datapoints) recorded at 100\,Hz has been downsampled to 50\,Hz.
It is split into 66 independent experiments as indicated by the discontinuities in the original time-series data.
Six out of 66 experiments have been utilized for testing whereas the other 60 experiments remain for training.
None of the reference methods from the model learning benchmark is out-of-the-box applicable to this large scale dataset.
To obtain a baseline, the sparse GP-NARX model has been trained on a subset of training experiments (400 inducing points, approx. 2000 training data points).
The PR-SSM can be directly trained on the full training dataset utilizing its stochastic, minibatched optimization scheme.
PR-SSM is setup similar to the configuration described in the benchmark experiment but based on a 14 dimensional latent state ($D_x = 14$).
Long-term prediction results on one of the test experiments are visualized in Fig.\,\ref{fig:LargeScaleResultsDetails}.
PR-SSM robustly predicts the robot arm motions for all joints and clearly improves over the GP-NARX baseline.
In contrast, the GP-NARX baseline can not predict the dynamics on 5 out of 7 joints.
\end{appendix}

\end{document}